%% file: arxiv-v2.tex
\documentclass[11pt,a4paper]{article}
\usepackage[hyperref]{acl2021}

\input{math_commands.tex}

\usepackage[ruled,vlined]{algorithm2e}
\usepackage{times}
\usepackage{latexsym}
\usepackage{amsmath}

\usepackage{graphicx}
\usepackage{hyperref}
\usepackage{url}
\usepackage{adjustbox}
\usepackage{enumitem}

\usepackage{amsfonts}

\usepackage[T1]{fontenc}
\usepackage[utf8]{inputenc}

\usepackage{microtype}

\aclfinalcopy 

\title{Scalable Transformers for Neural Machine Translation}

\author{Peng Gao$^{\dag}$, Shijie Geng$^{\ddag}$, Yu Qiao$^{\diamond}$, Xiaogang Wang$^{\dag}$, Jifeng Dai$^{\S}$, Hongsheng Li$^{\dag}$ \\
$^{\dag}$Multimedia Laboratory, The Chinese University of Hong Kong \ \ \  $^{\ddag}$Rutgers University \\
$^{\diamond}$Shenzhen Institute of Advanced Technology, Chinese Academy of Sciences \ \ \  $^{\S}$SenseTime Research \\
\texttt{1155102382@link.cuhk.edu.hk} \ \ \ \texttt{sg1309@rutgers.edu} \\
\texttt{yu.qiao@siat.ac.cn} \ \ \ \texttt{daijifeng@sensetime.com} \\
\texttt{\{xgwang, hsli\}@ee.cuhk.edu.hk} \\
}

\begin{document}
\maketitle
\begin{abstract}
Transformer has been widely adopted in Neural Machine Translation (NMT) because of its large capacity and parallel training of sequence generation. However, the deployment of Transformer is challenging because different scenarios require models of different complexities and scales. Naively training multiple Transformers is redundant in terms of both computation and memory.
In this paper, we propose a novel Scalable Transformers, which naturally contains sub-Transformers of different scales and have shared parameters. Each sub-Transformer can be easily obtained by cropping the parameters of the largest Transformer. A three-stage training scheme is proposed to tackle the difficulty of training the Scalable Transformers, which introduces additional supervisions from word-level and sequence-level self-distillation. Extensive experiments were conducted on WMT EN-De and En-Fr to validate our proposed Scalable Transformers.
\end{abstract}

\section{Introduction}

Transformers~\cite{vaswani2017attention} have demonstrated its superior performance on Machine Translation (NMT)~\cite{wu2016google,vaswani2017attention}, Language Understanding~\cite{devlin2018bert,brown2020language}, Image Recognition~\cite{touvron2020training}, and Visual-linguistic Reasoning~\cite{gao2019dynamic,lu2019vilbert,gao2019multi}. However, the scale (number of parameters and FLOPs) of the Transformers cannot be altered once trained. This is contradictory to the different scenarios of NMT which need models of different scales. For instance, NMT systems on a smartphone should have lower computational cost while those on clusters aim to achieve higher accuracy. A naive approach would be to separately train models of different scales. The Transformer of desired scale is then deployed to the target scenario. However, such a strategy requires training and storing multiple Transformers. A natural question is: can we build a single but Scalable Transformers that can be flexibly scaled up or down without re-training to run at different FLOPs, without sacrificing the translation accuracy at corresponding FLOPs? 

\begin{figure}[!t]
\includegraphics[width=\linewidth]{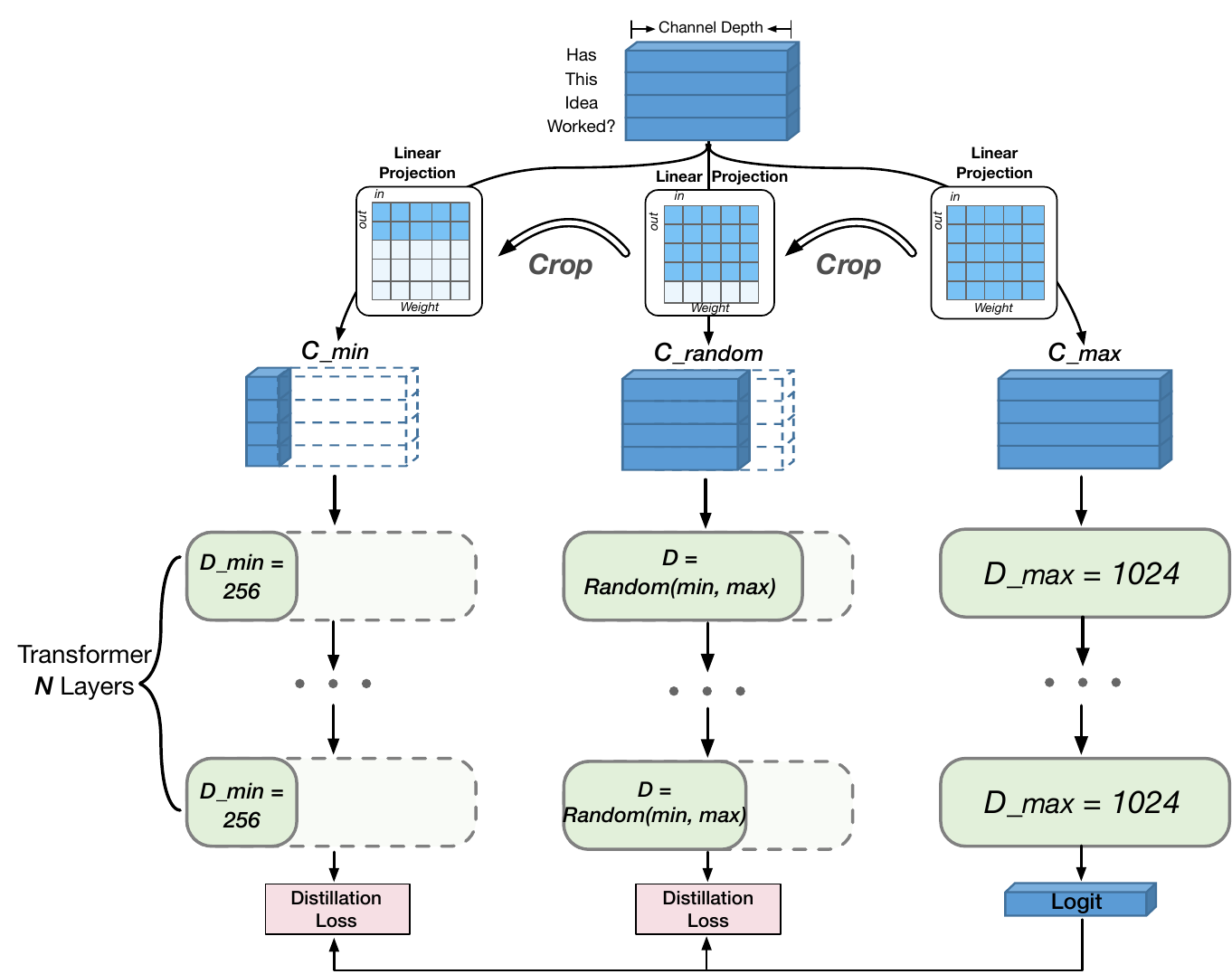}
\caption{Illustration of our proposed Scalable Transformer. After trained by the three-stage self-distillation training scheme, the Scalable Transformer contains a series of weight-sharing sub-Transformers for machine translation. The sub-Transformers with lower FLOPs can be easily obtained by cropping out from the widest Transformer without the need of re-training, thus saving redundant training and memory storage.}
\label{small_graph}
\end{figure}

In this paper, we propose a Scalable Transformers (ST) that can adjust feature dimensions of all encoder and decoder layers within a large range of widths (from 256 to 1024) without re-training. The largest model in our Scalable Transformers has 1024 feature dimensions at all layers, and 6 encoder and 6 decoder layers. After properly training, its neural layers can be cropped to form sub-models. For example, activating the first 512 dimension in each layer, we would obtain a sub-Transformer with $1/4$ of its full parameters but can still perform translation accurately with limited performance drop.
The sub-models of different scales share parameters with the largest model.

However, properly training the parameter-sharing sub-Transformers of different scales is non-trivial. Jointly training them results in worse performance than independently-trained counterparts, due to the interference between the parameter-sharing models. To solve the issue, we propose to incorporate online word-level and offline sequence-level self-distillation~\cite{kim2016sequence} to effectively supervise the training of the Scalable Transformers. During training, we randomly sample Transformers of different scales. The word-level predictions generated by the largest Transformer would serve as the additional supervisions for self-training smaller-scale sub-models with shared parameters. The training inference between sub-models can be mitigated by this strategy from two aspects. On the one hand, since the small sub-models share all their parameters with the largest Transformer, the predictions generated from the largest Transformer are easier to fit than the hard ground-truth supervisions. On the other hand, if the smaller sub-models are better trained, they would in turn enhance the performance of the largest model as their parameters are all included in the largest model. We then generate offline sequence-level predictions from the largest Transformer as supervisions for further finetuning the smaller-scale sub-models. After three stage training, all Transformers reach their optimal performances. Our Scalable Transformers have been tested on WMT'14 En-De and En-Fr benchmarks. All sub-models achieve better or comparable performances than their independently trained counterparts but with fewer overall parameters.


Our contributions can be summarized as threefold:
\begin{itemize}[leftmargin=*,noitemsep]
    \item We present a novel Scalable Transformers which can dynamically scale across a wide spectrum of parameters and FLOPs with shared parameters.
    
    \item We propose a novel three-stage training strategy for the Scalable Transformers, which include online word-level and offline sequence-level self-distillation for mitigating training interference between the sub-models.
    
    \item We perform extensive experiments on WMT'14 En-De and En-Fr datasets, which show favourable performances of our Scalable Transformers at different scales.
\end{itemize}

\section{Related Work}

\subsection{NMT Architectures}
Statistical Machine Translation (SMT) \cite{forcada2011apertium} dominated NMT in the early years. Seq2Seq \cite{sutskever2014sequence} then surpassed SMT and has become the mainstream for NMT. The main recent research turned to design architectures for seq2seq. NMT architectures evolved from LSTM \cite{hochreiter1997long}, CNN \cite{gehring2017convolutional}, DepthwiseCNN \cite{kaiser2017depthwise} to Transformer \cite{vaswani2017attention}. The Evolved Transformer \cite{so2019evolved} conducted architecture search using evolution algorithms and achieved good performance on NMT. Universal Transformer \cite{dehghani2018universal} proposed to share all layers of the encoder and decoder. Scaling Neural Machine Translation (SNMT) \cite{ott2018scaling} performs extensive ablation study on training the Transformer. Previous NMT research focuses on  architectures and training tricks while our Scalable Transformers is proposed to build an architecture that can be flexibly adjusted to meet a large spectrum of resource constraints with a single model.

\subsection{Distillation and Self-distillation}
Knowledge Distillation (KD)~\cite{hinton2015distilling} was first proposed to transfer knowledge between teacher and student networks. DistillBERT~\cite{sanh2019distilbert} adopts KD to train a small BERT model given a large BERT model. 
Sequence-level distillation~\cite{kim2016sequence} generalize KD to sequence generation. KD generally transfers knowledge between teacher and independent students different capacities.
There are also self-distillation methods \cite{yang2019snapshot,xie2019self,yang2019training,geng2021romebert} that use its previous versions to provide additional supervisions for improving itself. In contrast, we generate both online soft and offline hard supervisions from the largest model to train its parameter-sharing sub-models.

\subsection{Adaptive Neural Network}
Tradition neural networks have fixed architectures and computation complexity in training and testing. Recently, neural networks with  dynamic computation scales have been proposed in NLP and computer vision.
\citet{huang2016deep} proposed a stochastic depth training strategy for convolution neural networks, where residual layers are randomly dropped.
~\citet{huang2017multi,graves2016adaptive,figurnov2017spatially,graves2016adaptive} proposed adaptive computational mechanism for RNN, CNN and Transformer. For adaptive computation, the calculation would stop if the confidence score at certain layer is higher than a threshold. Adaptive computation can therefore early terminate the calculation for saving cost on easy-to-predict samples. ~\citet{gao2020multi} iteratively cascade weight-shared transformer for deeper neural machine translation. Slimmable Neural Network~\cite{yu2018slimmable, yu2019universally} was proposed to train width adjustable Convolution Neural Network (CNN). Although both slimmable NN and our Scalable Transformers adjust layer width to achieve different model capacities. There are two key differences. (1) Slimmable NN can only support equal layer widths for all layers while our Scalable Transformers allows flexibly setting different widths for different layers. (2) We handle training interference via a novel self-distillation training strategy while slimmable NN uses a small set of separate BN parameters, which are not feasible for our Scalable Transformers with much freedom. 

\subsection{Parameter Sharing}
Parameter sharing between a network and its sub-networks~\cite{pham2018efficient} has been
investigated for Neural Architecture Search (NAS). Different from~\citet{pham2018efficient}, we aim to train all sub-network jointly with the largest network and achieve good performance with self-distillation.

\section{Scalable Transformers}
\begin{figure*}[!tb]
\centering
\includegraphics[width=0.98\linewidth, keepaspectratio]{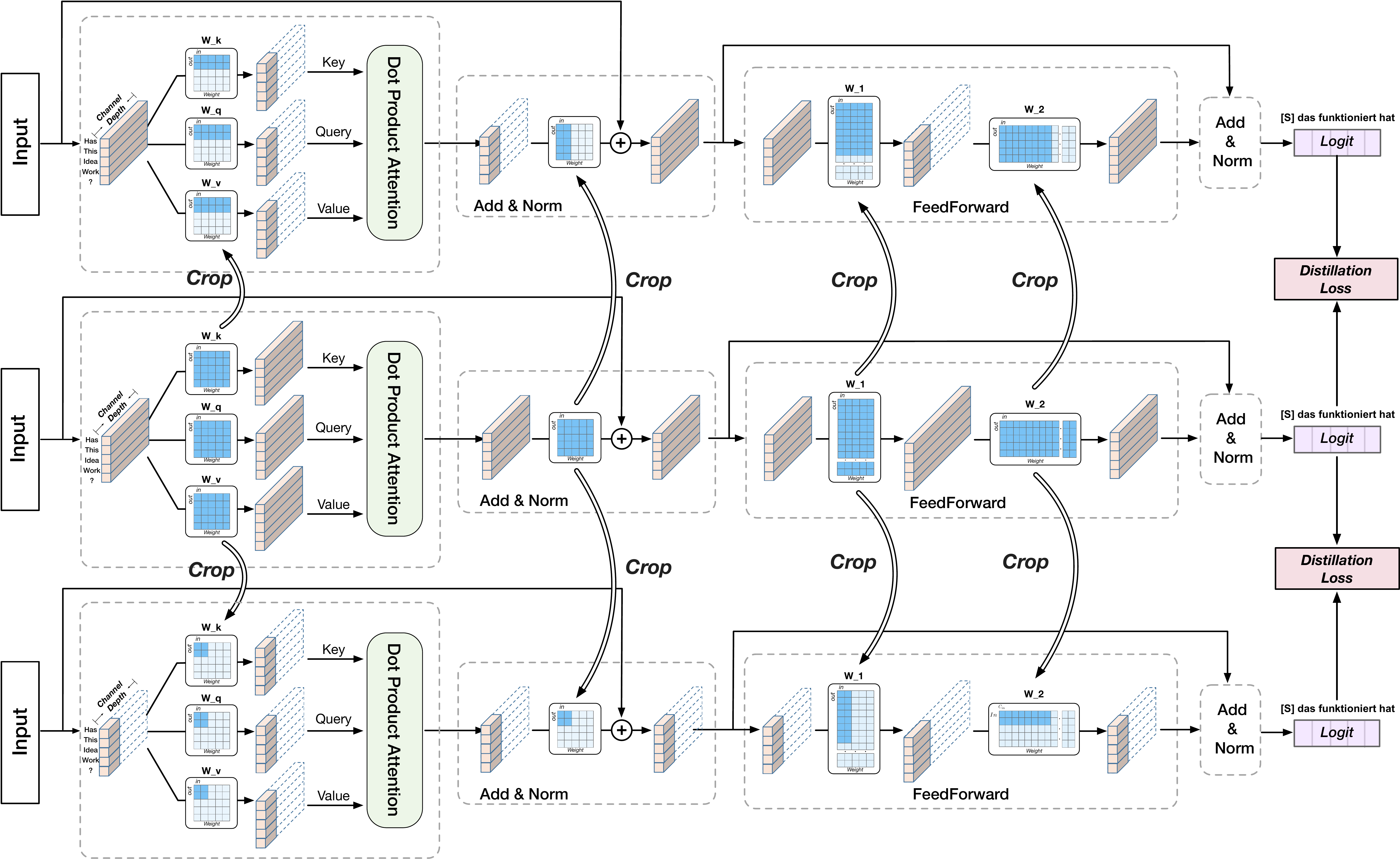}
\caption{The proposed Scalable Transformers (ST) can flexibly change different layers' feature width to meet different computational constraints. (Middle) The widest Transformer of a Scalable Transformers. (Top) A narrower sub-Transformer with fixed input-output feature dimension and varying attention feature dimensions for different layers. The weights of sub-Transformers are directly cropped from the widest Transformer without re-training. (Bottom) A narrower sub-Transformer with the same input-output and attention feature dimension for all layers.}
\label{overall}
\end{figure*}

In this section, we first briefly revisit the Transformer for machine translation. We then explain how to modify conventional Transformer to implement our Scalable Transformers (ST), which can flexibly choose its feature dimensions to match different target scales once trained (see Figure \ref{overall} for illustration). Since the training of the Scalable Transformers is non-trivial, we propose a three-stage training strategy, which consists of independent training (stage 1), online word-level self-distillation (stage 2), offline sequence-level self-distillation (stage 3).

\subsection{A Revisit of Transformer for Machine Translation}

\noindent \textbf{Embedding.}
Let $S \in \mathbb{R}^{N \times L}$ denote a source sentence of length $L$, where each word is represented by an one-hot vector of vocabulary size $N$. The input sentence $S$ are encoded by word embedding $W_e$ as $E = W_e S$, where $W_e \in \mathbb{R}^{N \times C}$.


\noindent \textbf{Transformer layer.}
Each Transformer layer contains one multi-head attention sub-layer followed by FFN sub-layer. The input features (word embedding $E$ for the 1st layer) are linearly projected into key, query, value, $K, Q, V \in \mathbb{R}^{L \times D}$ and propagated between positions as
\begin{align}
	K = &\,\, W_k E + b_k, \nonumber\\ 
	Q = &\,\, W_q E + b_q, \\
	V = &\,\, W_v E + b_v, \nonumber\\
	\mathrm{Attention}(Q,K,V) &= \operatorname{softmax} \left({Q K^{T}}/{\sqrt{d}}\right) V, \nonumber
\end{align}
where $W_k, W_q, W_v \in \mathbb{R}^{C\times D}$ and $b_k, b_q, b_v \in\mathbb{R}^D$. 
The multi-head attention further splits the $Q, K, V$ features along the channel dimension to form groups of features. Another linear projection converts the attention output feature back to $C$-dimensional followed by a two-layer FFN sub-layer,
\begin{align}
	\mathrm{FFN} = \mathrm{ReLU}(XW_1 + b_1)W_2 + b_2,
\end{align}
where $X$ is the input feature of the FNN, $W_1 \in \mathbb{R}^{C \times 4D}, b_1 \in \mathbb{R}^{4D}, W_2 \in \mathbb{R}^{4D \times C}, b_2 \in \mathbb{R}^C$ are the transformation parameters. Each Transformer layer contains the above two sub-layers.  A residual connection is also added around each of the above sub-layers followed by layer normalization. 

\subsection{Scalable Transformers}

\noindent \textbf{Architecture and word predictions.} The Transformer adopts an encoder-decoder architecture, each of which consists of 6 Transformer layers. Given the last output feature $O \in \mathbb{R}^{N\times C}$, it is mapped back to the word embedding space to predict the output words as $\mathrm{softmax} (O W_e^T)$, where $W_e$ is the transposed word embedding matrix.

The goal of our Scalable Transformers is that, once trained, the FLOPs and number of parameters of the Transformer can be flexibly adjusted  according to different use scenarios and without network re-training. To achieve this goal, we make the \emph{width} of our encoder and decoder layers to be flexibly modified within a pre-defined range.
Our Scalable Transformers shares the parameters with sub-models of different widths. Specifically, for each layer, the wider Transformer contains all the parameters and computations of the sub-Transformers. If the widest Transformer is properly trained, we can obtain smaller scale sub-Transformers by simply truncating Transformer layers' width and cropping the parameter matrices. Sub-Transformers cropped from the widest Transformer can still conduct accurate translation with limited performance drops. The widest Transformer layer has $C = M_{\max}$ and $D^i=M_{\max}$ as the input-output and attention feature dimension. The parameter matrices' sizes can be determined accordingly. For instance, we denote the attention matrix parameters at layer $i$ of the widest Transformer as $W^i_{k,\max}, W^i_{q, \max}, W^i_{v, \max}$, which are all of size $M_{\max} \times M_{\max}$. 

\noindent \textbf{Narrower Sub-Transformer layers.} Once the widest Transformer's parameters are defined, a narrower sub-Transformer's layer $i$ can be obtained by cropping a subset of parameters from the widest layer $i$ defined above. 
A narrower layer $i$ would adopt a smaller input-output dimension $C < M_{\max}$ as well as a smaller attention feature dimension $D^i < M_{\max}$. Without loss of generality, here we only discuss how to obtain the attention-key parameters of the narrower layer $i$, $W_k^i \in \mathbb{R}^{C\times D^i}$ and $b \in \mathbb{R}^D$. The same operations can be generalized to obtain other parameters ($W_q^i$, $W_v^i$,$W_o^i$ ,$W_1^i$, $W_2^i$, etc.) of the narrower Transformer's layer $i$. The attention-key parameters can be obtained as
\begin{align}
	W_k^i = W^i_{k,\max}[1\text{:}C, 1\text{:}D^i], \,\, b_k^i = b^i_{k,\max}[1\text{:}D^i], \nonumber
\end{align}
where $W^i_{k,\max}[1\text{:}C, 1\text{:}D^i]$ is the $C$-row and $D^i$-column top-left sub-matrix of the widest matrix $W_k^i$, and $b_k^i[1\text{:}D^i]$ denotes the first $D^i$ dimensions of the widest bias vector $b_k^i$. It is obvious that the widest Transformer layers contains all the parameters and computations of the narrower ones, and thus avoid introducing redundant parameters. The narrower sub-Transformers can then be obtained by stacking such sub-Transformer layers with different $D^i$ dimensions. Note that although the input-output dimension $C$ can be flexibly adjusted, we make the entire narrower sub-Transformer share the same $C$ dimensions across all layers because of the requirement of residual connections, while $D^i$ can be different for different layers. 

\noindent \textbf{Input and output projections.} Our Scalable Transformers only modifies the width of the Transformer layers. The dimension of the word embeddings remains to be $M_{\max}$ for the  sub-Transformers. To maintain the word embedding dimensions, for the widest Transformer, we use one additional linear projection to convert the word embeddings $E$ to 
\begin{align}
E' = W_{e, \max}' E + b'_{e,\max},
\end{align}
where $W_{e, \, \max}' \in \mathbb{R}^{M_{\max}\times M_{\max}}, b'_{e,\max} \in \mathbb{R}^{M_{\max}}$. Similarly, another linear projection is introduced to transform the FFN output features 
$O \in \mathbb{R}^{N \times M_{\max}}$ to 
\begin{align}
O' = W_{o,\max}'O + b_{o,\max}',
\end{align}
which are then used for final word predictions. For narrower sub-Transformers with smaller intermediate feature width $C < M_{\max}$, we crop the parameters from the widest input and output projections for sub-Transformers so that the word embedding can be shared across different scales,
\begin{align}
	W_e' &= W'_{e, \, \max}[1\text{:}M_{\max}, 1\text{:}C],\quad b_e' = b_{e,\max}'[1\text{:}C], \nonumber\\
	W_o' &= W'_{e, \, \max}[1\text{:}C, 1\text{:}M_{\max}], \quad b_o' = b_{o,\max}'.
\end{align}

\noindent \textbf{Scalable Transformer variants.} In our experiments, we mainly experiment with two types of Scalable Transformers. In the \emph{type-1} Scalable Transformers, each sub-Transformer's all layers adopt the same width from $\cal M$ for the input-output and attention feature dimensions, i.e., $C = D^1 = \cdots = D^{12}$. Therefore, there would be a total number of $|{\cal M}|$ sub-Transformers of different widths after the Scalable Transformers is trained. In the \emph{type-2} Scalable Transformers, we fix the input-output dimension $C = M_{\max}$ but allows freely choosing the attention dimension $D^i$ from any width in ${\cal M} = \{{M_1, \cdots, M_{\max}}\}$ at each layer. Therefore, a total number of ${|{\cal M}|}^{12}$ different sub-Transformers exist after the ST is trained.

\subsection{Training Scalable Transformers with Self-distillation} 
\label{sec:training}

Training the Scalable Transformers is quite challenging as it requires all sub-Transformers with shared parameters to have superior performance. Simply training all the sub-Transformers independently cannot result in satisfactory performances because of the gradient interference between sub-models. To tackle this issue, we propose a novel three-stage training strategy. The key idea of training is to distill the knowledge from the widest Transformer and use them as the supervisions for narrower sub-Transformers. Compared with ground-truth annotations, the predictions by the widest Transformers are easier to fit for the sub-Transformers as all of them share parameters to certain degrees. On the other hand, if the sub-Transformers are properly trained, their parameters could also boost the performance of wider Transformers that fully contain them. We propose a three-stage training strategy, where stage-1 focuses on training the widest Transformer and its sub-Transformers independently, and stage-2 and stage-3 utilize word-level and sequence-level self-distillation.

\noindent {\bf Stage 1: Joint sub-Transformer pre-training.} 
The stage-1 pre-training conducts jointly training on all the sub-Transformers. At each iteration $j$, we randomly sample a few sub-structures and always include the widest Transformer $T_{\max}$ for parameter updating. For each sub-Transformer, we randomly sample the input-output dimension $C$ and the attention width $D^i$ from $\{M_1, \cdots, M_{\max}\}$ for their Transformer layers following the layer width constraints of type-1 or type-2 models, i.e., we randomly sample from $|\cal M|$ and $|{\cal M}|^{12}$ sub-structures respectively for independent pre-training. All the sampled architectures are trained with the cross-entropy loss ${\cal L}_{ce}$ on ground truth words,
\begin{align}
	{\cal L}_1 = {\cal L}_{ce}(T_{\max}(S); G) + \sum_{T \in \mathbb{T}^{(j)}} {\cal L}_{ce}(T(S);G), 
\end{align}
where $G$ is the ground-truth sentence, $T_{\max}(S)$ and $T(S)$ denote the predicted words by the widest Transformer $T_{\max}$ and the sub-Transformer $T$ given the input sequence $S$, and $\mathbb{T}^{(i)}$ denotes the set of sampled sub-Transformers at iteration $j$.

Although joint sub-model pre-training actually results in worse performance than training only the widest Transformer, it pretrains all the sub-structures to have gradients of similar scales and also makes compromises on the shared parameters. Our stage-1 pre-training is almost the same as slimmable NN~\cite{yu2018slimmable, yu2019universally} because Transformer uses Layer Normliazation to avoid Batch Normalization. If removing the separate BNs in slimmable NN, it is equivalent to our stage-1, which jointly trains the largest model and its sub-models.

\begin{table*}
\centering
\scalebox{0.75}{
\begin{tabular}{lrrrrrrrrrrrrrl}
\hline \textbf{Feat. Dim.} & \textbf{256} & \textbf{320} & \textbf{384} & \textbf{448} & \textbf{512} & \textbf{576} & \textbf{640} & \textbf{704} & \textbf{768} &\textbf{832} &\textbf{896} &\textbf{960} &\textbf{1024}\\ \hline
Dropout & 0& 0& 0& 0.1& 0.1& 0.1& 0.1& 0.2& 0.2& 0.2& 0.3& 0.3& 0.3\\ \hline
\multicolumn{14}{c}{\quad \quad \quad With Input-output Projections} \\ \hline
\hline
Indep.   &  23.32 &  &  &  & 25.54 &  & & & 26.14 &   & & & 26.74  \\
Indep.+sep. teach   & 25.49  &  &  &  & 25.81 &  & & & 26.02 &   & & & 26.75  \\
Scalable+sep. teach & 25.53  & 25.49  & 25.71  & 25.78 & 25.99 & 26.02 & 26.05& 26.01 & 26.19 & 26.21  & 26.19 & 26.33 & 26.59  \\
\hline
Stage 1 & 23.76 & 23.85 & 24.27 & 24.57 & 24.86 & 25.25 & 25.56 & 25.94 & 26.23 & 26.21 & 26.50 & 26.58 & 26.61  \\
Stage 1+2 & 24.76 & 24.93 & 24.93 & 25.35 & 25.61 & 25.58 & 25.84 & 26.00 & 26.34 & 26.37 & 26.45 & 26.53 & 26.65  \\
Stage 2+3  & 24.91 & 24.59 & 25.11 &  25.12 &  25.32 & 25.36 & 25.41 &  25.53  & 25.57 & 26.01 & 25.99 & 26.19 & 26.21  \\
Stage 1+3  & 25.41 & 25.29 & 25.85 &  25.89 &  26.01 & 25.99 & 26.11 &  26.23  & 26.31 & 26.45 & 26.44 & 26.41 & 26.59  \\
Stage 1+2+3 & 25.59 & 25.59 & 26.06 & 26.05 & 26.15 & 26.11 & 26.06 & 26.23 & 26.30 & 26.49 & 26.61 & 26.65 & 26.71  \\
\hline
Params (M) & 45 & 51 & 59 & 68 & 79 & 91 &  104 & 119 & 135 & 152 & 171 & 191 & 209 \\ 
FLOPs (G) & 5.2 & 6.6 & 8.1 & 9.7 & 11.3 & 12.9 & 14.6 & 16.4 & 18.2 & 20.1 & 22.0 & 23.98 & 26.02 \\
\hline \hline
\multicolumn{14}{c}{\quad \quad \quad Without Input-output Projections} \\ \hline
Indep. &  23.35 &  &  &  & 25.56 &  & & & 26.12 &   & & & 26.73  \\
I/O Cropping & 25.11 & 25.07 & 25.18 & 25.42 & 25.71 & 26.02 & 25.98 & 26.07 & 26.32 & 26.22 & 26.31 & 26.48 & 26.46 \\
Params (M) & 19 & 28 & 37 & 48 & 61 & 75 & 90 & 106 & 124 & 144 & 163 & 186 & 209 \\
FLOPs (G) & 5.2 & 6.6 & 8.7 & 9.7 & 11.3 & 12.9 & 14.6 & 16.4 & 18.2 & 20.1 & 22.0 & 23.98 & 26.02 
\\
\hline
\end{tabular}
}
\caption{\label{newtest2013} Ablation study on En-De {\bf validation} (newtest2013) set with beam search 4. FLOPs are calculated with the assumption that source and target length are 20.}
\label{tab:ablation}
\end{table*}

\noindent {\bf Stage 2: Annealed word-level self-distillation.}
In stage-1 pre-training, the widest model is always updated in each iteration to ensure that it is more sufficiently trained than sub-Transformers. However, the interference between the models with shared parameters prevents them from effective learning. To mitigate the difficulty of training the sub-Transforms to fit the ``hard'' ground truth (one-hot vectors), we propose to distill the knowledge from the widest Transformer for training its sub-Transformers. Specifically, for each input sequence $S$, the soft predictions of the widest Transformer $T_{\max}(S)$ is used as additional training supervisions for its sub-Transformers $T$ with the cross-entropy loss,
\begin{align}
	{\cal L}_2 = &{\cal L}_{ce}(T_{\max}(S);\, G)+  \lambda_2^{(j)} \sum_{T \in \mathbb{T}^{(j)}} {\cal L}_{ce}(T(S);\, G) 	\nonumber \\ 
	& + (1-\lambda_2^{(j)}) \sum_{T \in \mathbb{T}^{(j)}} {\cal L}_{ce}(T(S);\, T_{\max}(S)),
\end{align}
where the first two terms are the same as the stage-1 loss, ${\cal L}_{ce}(T(S)\, ; T_{\max}(S))$ denotes using the widest model's predictions as learning targets for its sub-Transformers $T$, and $\lambda_2^{(j)}$ is a relative weight at iteration $j$ balancing the soft supervisions. 

In early iterations, our pre-trained Scalable Transformers from stage-1 still struggles because the predicted word probabilities $T_{\max}(S)$ might be noisy. Therefore, we make $\lambda_2^{(j)}$ closer to 1 when $j$ is small to more rely on the hard ground-truth words. $\lambda^{(j)}$ then gradually decreases to involve the soft predictions as additional supervisions as
\begin{align}
	\lambda_2^{(j)} = \begin{cases} 
	1 - 0.5\frac{j}{t_j}, & \text{ if } j < t_{j},\\
	0.5, & \text{ if } j \geq t_j.
	\end{cases}
\end{align}
$t_j$ is an iteration threshold, after which $\lambda_2^{(j)}$ is fixed to be $0.5$.
In this way, the sub-Transformers can gradually fit the soft predictions. Because the sub-Transformers' are also parts of the widest Transformer. The widest Transformer can also be improved, which in turn further provides more accurate supervisions for training sub-Transformers.

\noindent {\bf Stage 3: Sequence-level self-distillation.}
After stage-2 training, both the ground-truth labels as well as the word-level soft predictions might still be noisy to hinder the further improvements of the models. For stage-3 training, we conduct offline beam search with the widest Transformer to generate refined sequence-level predictions for all the training sequences. 
For our widest Transformer, it is trained with only the sequence-level refined predictions; for its sub-Transformers, they are trained with both the offline beam-searched predictions and online word-level predictions with the below loss,
\begin{align}
	{\cal L}_3 &= {\cal L}_{ce}(T_{\max}(S) ;\, T_{\max}^{\text{beam}}(S)) \nonumber \\
	& + \lambda_3 \sum_{T \in \mathbb{T}^{(j)}} {\cal L}_{ce}(T(S);\, T^{\text{beam}}_{\max}(S)) \nonumber\\
	& + (1-\lambda_3) \sum_{T \in \mathbb{T}^{(j)}} {\cal L}_{ce}(T(S);\, T_{\max}(S)), 
\end{align}
where the 1st term is the loss for training the widest model, the 2nd and 3rd terms are for training the sub-Transformers with both offline sequence-level predictions $T^{\text{beam}}_{\max}(S)$ and online word-level predictions $T_{\max}(S)$, and $\lambda_3$ is for weighting the two loss terms and is empirically fixed to $0.1$.

\section{Experiments}
\subsection{Dataset and Experiment Setup}

{\bf Datasets.} We test the ST on WMT 2014 English-German (En-De) and English-French (En-Fr) datasets. We adopt the En-De pre-processed version by \citet{vaswani2017attention} for fair comparison with previous approaches. Ablation study and hyperparameter tuning are conducted on WMT En-De validation (newstest2013) set and tested on the test (newstest2014). In En-Fr, we follow the same hyper-parameters of En-De. All ablation studies and experiments are conducted with type-1 Scalable Transformers.
BPE~\citep{sennrich2015neural} with shared dictionary between source and target was adopted for tokenization of both datasets. 32k and 40k joint dictionaries are created for En-De and En-Fr tasks. We measure the translation with case sensitive BLEU. All experiments use beam search 4 and length penalty 0.6. Following~\citet{vaswani2017attention}, we apply compound splitting.

\noindent {\bf Implementation.} 
Our Scalable Transformers contains 6 encoder layers and 6 decoder layers. There are 13 different widths in total, i.e. ${\cal M} = \{256, 320, \cdots, 1024\}$. The attention head dimension is fixed to $64$. 
We set different dropout rates within ST according to the features' dimensions. For En-De, we gradually increase the dropout rate $0\rightarrow 0.3$ as the feature dimension $256\rightarrow 1024$. For En-Fr, since it has larger-scale training data, we set dropout rates to 0 for widths in $[256, 576]$ and to $0.1$ for widths in $[640, 1024]$. Our optimization follows \citet{ott2018scaling} and details are can be found in supplementary materials.
Before stage-3 training, for the input sequence, we conduct prediction by the widest Transformer with beam search 4 and length penalty 0.6 to generate distillation targets. The predicted sequences are refined to remove the those whose source/target length ratio$>20$ or length$>250$ words. Our proposed and other compared methods' performances are reported based on the ensemble of the last 10 training epochs, following the experimental setting of \citet{ott2018scaling}.

\begin{table*}
\centering
\scalebox{0.9}{
\begin{tabular}{lccrrrrrrrrrrrl}
\hline \textbf{Feat. Dim.} & \textbf{256} & \textbf{320} & \textbf{384} & \textbf{448} & \textbf{512} & \textbf{576} & \textbf{640} & \textbf{704} & \textbf{768} &\textbf{832} &\textbf{896} &\textbf{960} &\textbf{1024}\\ \hline
Dropout & 0& 0& 0& 0.1& 0.1& 0.1& 0.1& 0.2& 0.2& 0.2& 0.3& 0.3& 0.3\\ \hline
Indep.   &  23.3 &  &  &  & 27.5 &  & & & 27.7 &   & & & 29.21  \\
Stage 1 & 24.0 & 24.3 & 24.6 & 25.0 & 25.7 & 26.3 & 27.0 & 27.4 & 27.8 & 28.3 & 28.6 & 28.75 & 28.74  \\
Stage 1+2 & 25.1 & 25.5 & 25.8 & 26.3 & 26.6 & 27.1 & 27.6 & 27.8 & 28.2 & 28.4 & 28.8 & 28.89 & 28.99  \\
Stage 2+3 & 26.7 & 26.6 & 27.0 & 27.1 & 27.6 & 27.4 & 27.9 & 28.1 & 28.4 & 28.7 & 28.6 & 28.76 & 29.18  \\
Stage 1+3 &   26.9 &26.8 & 27.1 & 27.3 & 27.5 & 27.9 &  28.2 & 28.3 & 28.6 & 28.7 & 28.7 & 28.94 & 29.01  \\
Stage 1+2+3 & 27.1 & 27.3 & 27.5 & 27.9 & 28.3 & 28.4 & 28.6 & 28.7 & 29.0 & 28.9 & 29.2 & 29.34 & 29.27 \\ 
\hline
\end{tabular}
}
\caption{Performance of type-1 Scalable Transformers on En-De \textbf{test} (newtest2014) set.}
\label{newtest2014}
\end{table*}

\begin{table}
\centering
\small
\scalebox{0.91}{
\begin{tabular}{lcc}
\hline \textbf{Model} & \textbf{EN-De} & \textbf{En-Fr} \\ \hline
{Transformer} \citep{vaswani2017attention} & 28.4 & 40.5 \\
{Weighted Transformer} \citep{ahmed2017weighted}& 28.9 & 41.0 \\ 
{Relative Transformer} \citep{shaw2018self}& 29.2 & 41.5 \\ 
{Scaling NMT} \citep{ott2018scaling} & 29.3& 43.2 \\ \hline
{Evolved Transformer} \citep{so2019evolved} & 29.8 &  - \\ 
(NAS on 200 TPUs) \\ \hline
Baseline& 29.2 & 43.1\\
Type-1 Stage 1 (widest)  & 28.7 & 42.8\\
Type-1 Stage 1+2+3 (widest) & 29.3 & 43.1\\
Type-2 Stage 1+2+3 (widest) & 29.5 & 43.1 \\
Type-2 Stage 1+2+3 (sub-model) & 29.7 & 43.3 \\
\hline
\end{tabular}
}
\caption{\label{sota} Comparison between Scalable Transformers with state-of-the-art methods on the \textbf{test} sets.}
\label{tab:final}
\end{table}

\subsection{Ablation Study on WMT En-De}

\noindent {\bf Independent vs. Scalable Transformers.} 
We first independently train separate Transformers, which have unified feature widths of $\{256, 512, 768, 1024\}$ for all layers but have fixed $1024$-d word embedding and the additional input-output projections for fairly comparing with our type-1 Scalable Transformers' sub-Transformers. Our Scalable Transformers shows much better performance than the independently trained counterparts (see ``Stage 1+2+3'' vs ``Indep.'' in Table \ref{tab:ablation}).

\noindent {\bf Stage-1 vs. stage-2 vs. stage-3 training.} We then show the necessity of the proposed three-stage training scheme. Simple joint training all sub-Transformers in stage-1 (equivalent to slimmable NN \citep{yu2018slimmable} as we discussed in Sec. \ref{sec:training}) results in even worse performance than independent training (``Stage 1'' vs. ``Indep.'' in Table \ref{tab:ablation}), which illustrate the training interference between different sub-Transformers.
Stage-2 training shows significant improvements over stage-1 results. Stage-3 training further improves the performance with offline sequence-level self-distillation (``Stage 1+2'' and ``Stage 1+2+3'' in Table \ref{tab:ablation}). 
We also test discarding stage-1 training (``Stage 2+3'') and  discarding stage-2 training (``Stage 1+3'') from our three-stage training scheme. Both strategies lead to inferior performances compared with our proposed three-stage training scheme.

\noindent {\bf Independent models + separate widest teacher.} The above experiments show that parameter-sharing sub-models and the full model can mutually boost their performance via the proposed training scheme. To show the improvement is not from simple knowledge distillation, we train an independent widest Transformer to generate the refined sequence-level predictions. After the training converges, we use the widest transformer as the teacher outputting both online soft and offline beam-searched hard targets to guide the training of 12 independent models with varying widths 256-1024. Results show that our Scalable Transformer can achieve better performance than independent distillation (``Indep.+sep. teach'' vs. ``indep.'') due to the parameter sharing between models of different width, especially on narrower models.
(``Indep.+sep. teach'' vs. ``Stage 1+2+3'' in Table \ref{tab:ablation}).

\noindent {\bf Scalable Transformer + separate widest teacher.}
An alternative approach to the above study is to keep the separate widest Transformer as the teacher to teach a proposed Scalable Transformer. The performances by this strategy (``scalable+sep. teach'' vs. ``Stage 1+2+3'') are only slightly lower than our proposed self-distilled scalable model. The reason might be that, although the scalable models share parameters and are jointly trained. The separate widest teacher's outputs might be less compatible with those of the narrower models than the scalable model's widest Transformer. In addition, this training strategy requires more training costs. For each forward process, only 46G FLOPs are required for our proposed training scheme, while 72G FLOPs are needed for this separate widest teacher strategy, as the separate teacher model needs to be separately forwarded to obtain the word-level soft targets. Training the separate widest model also requires additional resources.


\noindent {\bf Advantages of type-2 ST.} Our type-2 ST can more flexibly adjust its FLOPs because it can freely choose different feature widths $D^i$ for different layers. Surprisingly, after the type-2 model is trained, we found many of its sub-Transformers can even surpass the widest Transformer. Given the widest Transformers of our trained type-2 model from the last 10 epochs, for each layer width $D^i$, we randomly choose the feature dimension from $\{896, 960, 1024\}$ and conduct evaluation with $1,000$ random such sub-Transformers on the En-De validation set. We choose the best sub-Transformer on the validation (newtest2013) set and evaluate on the test (newtest2014) set. Such an optimal sub-model achieves $26.9$ and $29.7$ on the validation and test sets (see Table~\ref{tab:en-de-beam} in supplemental materials), which are even higher than the widest Transformer. We also calculate the mean and standard deviation of the BLEU scores of the top-10 sub-Transformers, which achieve $26.82\pm0.036$ and $29.60\pm 0.061$ on validation and test sets. All top-10 sub-Transformers' performances are higher than those of the widest Transformer ($26.7$ for validation and $29.3$ for test), demonstrating the effectiveness of our type-2 Scalable Transformers. We will continue to investigate how to improve the efficiency on searching the optimal sub-model. To verify that the performance improvement of type-2 Transformer is not from the 1,000 searched sub-models, we store 1,000 checkpoints in the last 3 epochs when training an independent widest transformer. We average the best 10 models out of 1,000 checkpoints, which achieve $29.4$ on En-De test set and is very similar to the baseline's $29.3$. However, such a result is still worse than our searched optimal Type-2 sub-model's $29.7$.

\noindent {\bf Training time and memory of Scalable Transformer.}
Our proposed Scalable Transformer has 209M parameters and is trained for 212 hours on 8 V100 GPUs. If one separately trains 12 Transformers with varying width, it would use 1290M parameters and 712 hours for training, both of which are much larger than those of our proposed method.

\noindent {\bf Other factors.} We provide more ablation studies in the appendix.

\subsection{Final Results on WMT EN-De and En-Fr}

We test the proposed Scalable Transformers on WMT'14 EN-De test (newtest2014) set and En-Fr test set. Performances of type-1 ST on En-De test (newtest2014) are listed in Table~\ref{newtest2014}, which show similar tendency as those on En-De validation set.

We further compare the widest model from our type-1 Scalable Transformers and the randomly searched optimal sub-model from our type-2 Scalable Transformers with the original Transformer \citep{vaswani2017attention}, weighted Transformer~\citep{ahmed2017weighted}, relative position Transformer~\citep{shaw2018self}, scaling neural machine translation~\citep{ott2018scaling} and evolved Transformer \citep{so2019evolved} in Table~\ref{sota}. The baseline Transformer is based on~\citet{ott2018scaling}. Our reproduction achieves $29.2$ and $43.1$ on En-De and En-Fr, comparable with the original results. The simple joint training (denoted as ``Type-1 Stage-1 (widest)'') results in slightly worse performance than the independently trained baseline.

The widest Transformer from type-1 ST achieves $29.3$ and $43.1$ on En-De and En-Fr, which are comparable with the baseline Transformer, but with 13 sub-models of different FLOPs and parameter sizes. The best randomly searched sub-Transformer of our type-2 ST (denoted as ``Type-2 Stage 1+2+3 (sub-model)'') on En-De can even reach $29.7$, which has fewer parameters but outperforms state-of-the-art standard Transformers with $1024$-dimensional features. Evolved Transformer performed architecture search with 200 TPUs to find the optimal translation architecture. It achieved $29.8$ BLEU with 218M parameters. In contrast, our type-2 Scalable Transformers is much easier to implement with existing libraries, only used 3 days on 8 V100 GPUs for training, and achieved comparable $29.7$ BLEU with 221M parameters. 

\section{Conclusion}
In this paper, we propose a novel Scalable Transformers to achieve efficient and robust deployment of Transformers of different scales. We choose to make the Scalable Transformers change its scale by modifying its layer width. Through carefully designed word-level and sequence-level self-distillation, the proposed Scalable Transformers can be trained with an incremental increase of training time. After training, sub-Transformers of different scales can be easily obtained by cropping from the widest Transformer to achieve flexible deployment.

\bibliography{anthology,arxiv}
\bibliographystyle{acl_natbib}

\appendix
\section{Optimization Details} 

ADAM \citep{kingma2014adam} optimizer is adopted with $\beta_{1} = 0.9$, $\beta_{2} = 0.98, \epsilon = 10^{-8}$. Other optimization details can be found in Table \ref{tab:optimization}. 
The learning rate of each stage linearly increases for the first $4,000$ iterations from $5\times 10^{-4}$ to the maximum learning rate of each stage (denoted as ``max lr'' in Table \ref{tab:optimization}). After reaching the maximum learning rate, the learning rate decays according to 
\begin{align}
	 \mathrm{lr} = \frac{\mathrm{lr}_{\mathrm{max}} \times \sqrt{4000}}{\sqrt{\#\mathrm{iteration}}}
\end{align}
where $\mathrm{lr}_{\mathrm{max}}$ stands for the maximum, $4000$ is the number of iterations for warming up, $\#\mathrm{iteration}$ stands for the current iteration number. The learning rate scheme follows the original Transformer \citep{vaswani2017attention}. The total training epoch for each stage is denoted as ``Epoches'' in Table~\ref{tab:optimization}. Label smoothing of 0.1 is conducted following \citet{ott2018scaling}. Weight decay is not used for all experiments. We train the proposed Scalable Transformer with 8 V100 GPUs, each of which holds 3584 and 5120 tokens for En-De and En-Fr, respectively. We adopt gradient accumulation of 16 to further increase the training batch size. $\lambda_2$ or $\lambda_3$ weight the contributions of the cross-entropy loss and self-distillation loss and their settings are recorded in Table~\ref{tab:optimization} (denoted as ``$\lambda_2$/$\lambda_3$'').

\begin{table}
\centering
\begin{adjustbox}{width=0.85\linewidth}
\begin{tabular}{lccc}
\hline {Training stage} & {Epoches} &  {max lr} & {$\lambda_2$/$\lambda_3$} \\ \hline
EN-De stage-1 &  60 &  0.007& n/a   \\
En-De stage-2 &  60 &  0.007& 0.5  \\
En-De stage-3 &  30 &  0.004& 0.9  \\ \hline
En-Fr stage-1 &  30 &  0.006& n/a    \\
En-Fr stage-2 &  30 &  0.006& 0.5  \\
EN-Fr stage-3 &  15 &  0.004& 0.9  \\
\hline
\end{tabular}
\end{adjustbox}
\caption{The hyperparameters for optimization on En-De and En-Fr datasets.}
\label{tab:optimization}
\end{table}

\section{Additional Ablation Studies and Experiments}

\noindent {\bf Dropout scheme.} Dropout has major impact on the final performance for training Transformers. Our Scalable Transformer gradually increase the dropout rates from $0.1$ to $0.3$ as the feature dimensions increase from 256 to 1024. We test two alternative dropout strategies. The first one is to set dropout rates as $0.3$ for all feature dimensions in $\{256,\cdots, 1024\}$. Another strategy is to set dropout rate to 0 for all feature dimensions except for dimension 1024, which has a dropout rate of $0.3$.
The results on En-De validation set is shown in Table~\ref{tab:dropout}, which show that our proposed dropout strategy leads to better performance than those simplified ones on all widths of the type-1 Scalable Transformer.

\noindent {\bf Effects of input-output linear projections.} Our Scalable Transformers introduces two additional linear projections $W_o'$ and $W_e'$ to ensure that the same word embedding can be used across sub-Transformers. We show that the gain is not from the extra linear projections. We independently train another 4 Transformer to have the same word embedding and feature dimensions of $\{256, 512, 768, 1024\}$ respectively. Those 4 Transformers do not need the additional linear projections. The two ``Indep.'' in Table \ref{tab:ablation} show that they have almost the same performance as independent Transformers with projections. Another possible way to replace the input-output projections is to crop the top corner of the embedding matrix to obtain narrower word embedding for the sub-Transformers. This strategy (``I/O Cropping'' in Table \ref{tab:ablation}) show inferior accuracy compared with the proposed additional input-output linear projections.

\begin{table}
\centering
\begin{adjustbox}{width=1.0\linewidth}
\begin{tabular}{cccc}
\hline Encoder-decoder Layers & Layer Width & EN-De Test & Parameter (M)\\ \hline
6-6& 512 & 27.5 & 61\\
3-3& 768 & 26.3 & 73\\
\hline
\end{tabular}
\end{adjustbox}
\caption{Performance of (top) deep but thin Transformer versus (bottom) shallow but wide Transformer with similar parameter sizes on En-De \textbf{test} set. The former network has smaller performance drop.}
\label{depth_width}
\end{table}

\noindent {\bf Beam search.} We also test the impact of using beam search during inference. The results are listed in Tables \ref{tab:en-de-beam} and \ref{tab:en-fr-beam}. Beam search generally improves the translation accuracy, especially on stage-1 or stage-2 results. After stage-3 training, the improvements of ``Stage 1+2+3'' with beam search over ``Stage 1+2+3'' without beam search become marginal on the two datasets ($+0.17$ for En-De test set and $+0.04$ for En-Fr test set), which indicate that our three-stage training strategy can eliminate the need for beam search at inference to certain extent.

\begin{table*}
\centering
\scalebox{0.8}{
\begin{tabular}{lrrrrrrrrrrrrrl}
\hline \textbf{Feat. Dim.} & \textbf{256} & \textbf{320} & \textbf{384} & \textbf{448} & \textbf{512} & \textbf{576} & \textbf{640} & \textbf{704} & \textbf{768} &\textbf{832} &\textbf{896} &\textbf{960} &\textbf{1024}\\ \hline
\multicolumn{14}{c}{\quad \quad  Proposed Dropout Stratety} \\ \hline
Dropout & 0& 0& 0& 0.1& 0.1& 0.1& 0.1& 0.2& 0.2& 0.2& 0.3& 0.3& 0.3\\ \hline
Stage 1+2+3 & 25.59 & 25.59 & 26.06 & 26.05 & 26.15 & 26.11 & 26.06 & 26.23 & 26.30 & 26.49 & 26.61 & 26.65 & 26.71  \\
\hline \hline
\multicolumn{14}{c}{\quad \quad  Uniform Maximal Dropout} \\ \hline
Dropout & 0.3& 0.3& 0.3& 0.3& 0.3& 0.3& 0.3& 0.3& 0.3& 0.3& 0.3& 0.3& 0.3\\ \hline
Stage 1+2+3 & 23.87 & 23.97 & 24.28 & 24.32 & 24.15 & 24.32 & 24.69 & 25.74& 26.23 & 25.96 & 26.45 & 26.32 & 26.47  \\ \hline  \hline
\multicolumn{14}{c}{\quad \quad  Widest Dropout Only} \\ \hline
Dropout & 0& 0& 0& 0& 0& 0& 0& 0& 0& 0& 0& 0& 0.3\\ \hline
Stage 1+2+3 & 24.89 & 24.91 & 24.95 & 25.02 & 25.17 & 25.28 & 25.34 & 25.56 & 25.88 & 26.10 & 26.22 & 26.01 & 26.35  \\ \hline

\end{tabular}
}
\caption{\label{appendix_newtest2013} Tuning dropout rates on type-1 Scalable Transformer when training on En-De \textbf{validation} (newtest2013) set with beam search 4.}
\label{tab:dropout}
\end{table*}

\noindent \textbf{Width adaptive or depth adaptive?} In our proposed Scalable Transformer, we decide to tune layer width of the Transformer to achieve adjustable FLOPs and parameter sizes. To show tuning layer width is better than tuning network depth, we implement two independent Transformers with approximately the same number of parameters. One contains 6 encoder and 6 decoder layers with feature dimension 512. The other one contains 3 encoder and 3 decoder layers with feature dimension of 768. Their translation accuracy on En-De test (newtest2014) set is reported in Table~\ref{depth_width}, which show that decreasing the depth of the Transformer significantly impacts the translation accuracy $29.2\rightarrow 26.3$, while decreasing the layer width has smaller performance drop $29.2\rightarrow 27.4$. Therefore, it is better to tune the width of the Transformer layers than the overall depth of the Transformer to better maintain the translation accuracy.

\noindent \textbf{Top-10 randomly searched type-2 sub-Transformers.} 
We conduct random search on type-2 Scalable Transformer's sub-models to find the optimal performance as stated in Sec. 4.2 of the main paper. Table~\ref{search} shows the top-10 sub-models' layer widths and their performances on En-De validation and test sets. The results clearly show that many sub-Transformers of our type-2 ST can even surpass the widest Transformer, which worth further studying how to effectively conduct sub-model search in the future.

\begin{table*}
\centering
\scalebox{0.85}{
\begin{tabular}{lrrrrrrrrrrrrrl}
\hline \textbf{Feat. Dim.} & \textbf{256} & \textbf{320} & \textbf{384} & \textbf{448} & \textbf{512} & \textbf{576} & \textbf{640} & \textbf{704} & \textbf{768} &\textbf{832} &\textbf{896} &\textbf{960} &\textbf{1024}\\ \hline
Dropout & 0& 0& 0& 0.1& 0.1& 0.1& 0.1& 0.2& 0.2& 0.2& 0.3& 0.3& 0.3\\ \hline
\multicolumn{14}{c}{\quad \quad  Beam Search 4 } \\ \hline
Stage 1 & 24.0 & 24.3 & 24.6 & 25.0 & 25.7 & 26.3 & 27.0 & 27.4 & 27.8 & 28.3 & 28.6 & 28.75 & 28.74  \\
Stage 1+2 & 25.1 & 25.5 & 25.8 & 26.3 & 26.6 & 27.1 & 27.6 & 27.8 & 28.2 & 28.4 & 28.8 & 28.89 & 28.99  \\
Stage 1+2+3 & 27.1 & 27.3 & 27.5 & 27.9 & 28.3 & 28.4 & 28.6 & 28.7 & 29.0 & 28.9 & 29.2 & 29.34 & 29.27  \\
\hline
\multicolumn{14}{c}{\quad \quad  Beam Search 1 } \\ \hline
Stage 1 & 23.3 & 23.4 & 23.7 & 24.3 & 24.8 & 25.5 & 26.1 & 26.9 & 27.2 & 27.2 & 27.3 & 27.58& 27.90  \\
Stage 1+2 & 23.7 & 24.2 & 24.6 & 25.1 & 25.5 & 26.0 & 26.7 & 27.0 & 27.2 & 27.4 & 27.6 & 28.02& 28.19  \\
Stage 1+2+3 & 26.7 & 26.8 & 27.1 & 27.4 & 27.8 & 28.0 & 28.1 & 28.3 & 28.6 & 28.7 & 28.8 & 28.89& 29.10  \\ \hline
Params (M) & 46 & 52 & 60 & 69 & 79 & 91 &  104 & 119 & 135 & 152 & 171 & 191 & 209 \\ 
FLOPs (G) & 5.2 & 6.6 & 8.1 & 9.7 & 11.3 & 12.9 & 14.6 & 16.4 & 18.2 & 20.1 & 22.0 & 23.98 & 26.02 \\
\hline
\end{tabular}
}
\caption{\label{en-de} Performance of type-1 Scalable Transformer on En-De \textbf{test} (newtest2014) set with beam search 4 and 1.}
\label{tab:en-de-beam}
\end{table*}

\begin{table*}
\centering
\scalebox{0.85}{
\begin{tabular}{lrrrrrrrrrrrrrl}
\hline \textbf{Feat. Dim.} & \textbf{256} & \textbf{320} & \textbf{384} & \textbf{448} & \textbf{512} & \textbf{576} & \textbf{640} & \textbf{704} & \textbf{768} &\textbf{832} &\textbf{896} &\textbf{960} &\textbf{1024}\\ \hline
Dropout & 0 & 0& 0& 0& 0& 0& 0.1&0.1 & 0.1& 0.1& 0.1 & 0.1& 0.1\\ \hline
\multicolumn{14}{c}{\quad \quad  Beam Search 4 } \\ \hline
Indep.  &    &  &  &  &  & & & &  & & & & 43.11  \\ 
Stage 1 & 39.1 & 38.9 & 39.2 & 39.6 & 39.9 & 40.5 & 40.9 & 41.4 & 41.8 & 42.0 & 42.3 & 42.41 & 42.84  \\
Stage 1+2 & 39.4 & 39.5 & 40.0 & 40.2 & 40.6 & 41.0 & 41.3 & 41.9  & 42.2 & 42.4 & 42.6 & 42.65 & 42.95  \\
Stage 1+2+3 & 40.9 & 41.0 & 41.3 & 41.5 & 41.7 & 41.7 & 42.0 & 42.2 & 42.4 & 42.5 & 42.9 & 43.03 & 43.07  \\ \hline
\multicolumn{14}{c}{\quad \quad  Beam Search 1 } \\ \hline
Stage 1 & 38.0 & 38.0 & 38.5 & 38.9 & 39.3 & 39.7 & 40.5 & 40.8 & 41.3 & 41.5 & 41.7 & 41.92 & 42.11  \\
Stage 1+2 & 38.6 & 38.6 & 39.1 & 39.4 & 39.8 & 40.2 & 40.7 & 41.1 & 41.4 & 41.7 & 41.9 & 42.10 & 42.40  \\
Stage 1+2+3 & 40.4 & 40.6 & 40.8 & 41.0 & 41.4 & 41.5 & 41.9 & 42.1 & 42.3 & 42.4 & 42.7 & 42.84 & 43.03  \\ \hline
Params (M) & 57 & 63 & 71 & 80 & 91 & 103 & 116 & 130 & 146 & 164 & 182 & 202 & 224 \\   
FLOPs (G) & 5.2 & 6.6 & 8.1 & 9.7 & 11.3 & 12.9 & 14.6 & 16.4 & 18.2 & 20.1 & 22.0 & 23.98 & 26.02 \\
\hline
\end{tabular}
}
\caption{\label{en-fr} Performance of type-1 Scalable Transformer on En-Fr \textbf{test} set with beam search 4 and 1. }
\label{tab:en-fr-beam}
\end{table*}


\begin{table*}
\centering
\scalebox{0.8}{
\begin{tabular}{lccc}
\hline \textbf{Model} & \textbf{newtest 2013} & \textbf{newtest 2014} & FLOPs(G)\\ \hline
\textbf{Widest:} $[1024, 1024, 1024, 1024, 1024, 1024, 1024, 1024, 1024, 1024, 1024, 1024]$ & 26.73 &29.44 &26.02  \\ \hline \hline
$[1024, 1024, 1024, 1024, 896, 1024, 1024, 960, 1024, 1024, 1024, 896]$ & 26.87 &29.66 &25.30  \\
$[1024, 1024, 1024, 1024, 960, 896, 1024, 896, 960, 1024, 960, 960]$ & 26.85  &29.64 & 24.86 \\
$[1024, 1024, 1024, 1024, 960, 896, 1024, 896, 960, 1024, 1024, 896]$ & 26.85  &29.62 & 24.86  \\
$[1024, 1024, 1024, 1024, 960, 1024, 1024, 896, 1024, 1024, 960, 896 ]$ & 26.83 &29.62 & 25.06 \\
$[1024, 1024, 1024, 1024, 960, 896, 1024, 896, 960, 1024, 960, 896 ]$ & 26.83  &29.60 & 24.69 \\
$[1024, 1024, 1024, 1024, 1024, 1024, 1024, 896, 1024, 1024, 1024, 960]$ & 26.82 &29.58 &25.15 \\
$[1024, 1024, 1024, 1024, 1024, 1024, 1024,  960, 1024, 1024, 960, 960]$ & 26.80 & 29.54 & 25.50 \\
$[1024, 1024, 1024, 1024, 960, 1024, 1024, 1024, 1024, 1024, 1024, 896]$ & 26.77 & 29.69 & 25.57\\
$[1024, 1024, 1024, 1024, 1024, 1024, 1024,  1024, 1024, 960, 1024, 896]$ & 26.77 &29.47 & 25.50 \\
$[1024, 1024, 1024, 1024, 1024, 1024, 1024, 896, 1024, 1024, 1024, 960]$ & 26.76  &29.64 & 25.50 \\
\hline
\end{tabular}
}
\caption{BLEU scores of the top-10 randomly searched sub-Transformers from our type-2 Scalable Transformer on En-De \textbf{validation} (newtest2013) and \textbf{test} (newtest2014) sets.}
\label{search} 
\end{table*}

\end{document}

%% file: math_commands.tex
\usepackage{amsmath,amsfonts,bm}

\def\eqref#1{equation~\ref{#1}}

\def\1{\bm{1}}

\DeclareMathAlphabet{\mathsfit}{\encodingdefault}{\sfdefault}{m}{sl}
\SetMathAlphabet{\mathsfit}{bold}{\encodingdefault}{\sfdefault}{bx}{n}

